\pdfoutput=1

\documentclass[11pt]{article}

\usepackage{acl}

\usepackage{times}
\usepackage{latexsym}

\usepackage[T1]{fontenc}

\usepackage[utf8]{inputenc}
\usepackage{fancyhdr}
\usepackage{microtype}
\usepackage{natbib}
\usepackage{csquotes}
\usepackage{booktabs}
\usepackage{amsmath}
\usepackage{graphicx, subcaption, wrapfig, tabularx, listings, array, enumitem, csquotes, ,makecell, float}
\usepackage{algorithm}
\usepackage{algpseudocode}

\title{Werewolf Arena: A Case Study in LLM Evaluation via Social Deduction}

\author{Suma Bailis\thanks{*Corresponding author}, Jane Friedhoff, \and  Feiyang Chen \\
        Google Research \\ 
        \texttt{\{sbailis,jfriedhoff,feiyangc\}@google.com}}

\begin{document}
\pagestyle{fancy}
\fancyhf{} 
\renewcommand{\headrulewidth}{0pt} 
\cfoot{\thepage} 
\maketitle
\begin{abstract}
This paper introduces Werewolf Arena, a novel framework for evaluating large language models (LLMs) through the lens of the classic social deduction game, \textit{Werewolf}. In Werewolf Arena, LLMs compete against each other, navigating the game's complex dynamics of deception, deduction, and persuasion.  The framework introduces a dynamic turn-taking system based on bidding, mirroring real-world discussions where individuals strategically choose when to speak. We demonstrate the framework's utility through an arena-style tournament featuring Gemini and GPT models. Our results reveal distinct strengths and weaknesses in the models' strategic reasoning and communication. These findings highlight Werewolf Arena's potential as a challenging and scalable LLM benchmark.
\end{abstract}

\section{Introduction}
Creating truly human-like AI requires sophisticated cognitive abilities such as reasoning about others' intentions, navigating deceptive information, and convincingly communicating in complex social settings. Evaluating these nuanced skills in both humans and LLMs poses a significant challenge, as traditional benchmarks often fall short \cite{Ullman2023LLMTheoryOfMind}.

Social Deduction Games (SDGs), such as the popular game \textit{Werewolf}, present a compelling avenue for addressing this challenge. SDGs encapsulate many of the complexities of human social interaction, requiring players to utilize various reasoning skills (e.g., temporal, deductive, and inductive) within an uncertain environment. In \textit{Werewolf}, Villager and Werewolf players engage in a battle of wits, leveraging deception and persuasion to achieve their respective goals.

\begin{figure*}[ht!]
\begin{minipage}{\textwidth}
\centering
\includegraphics[scale=0.26]{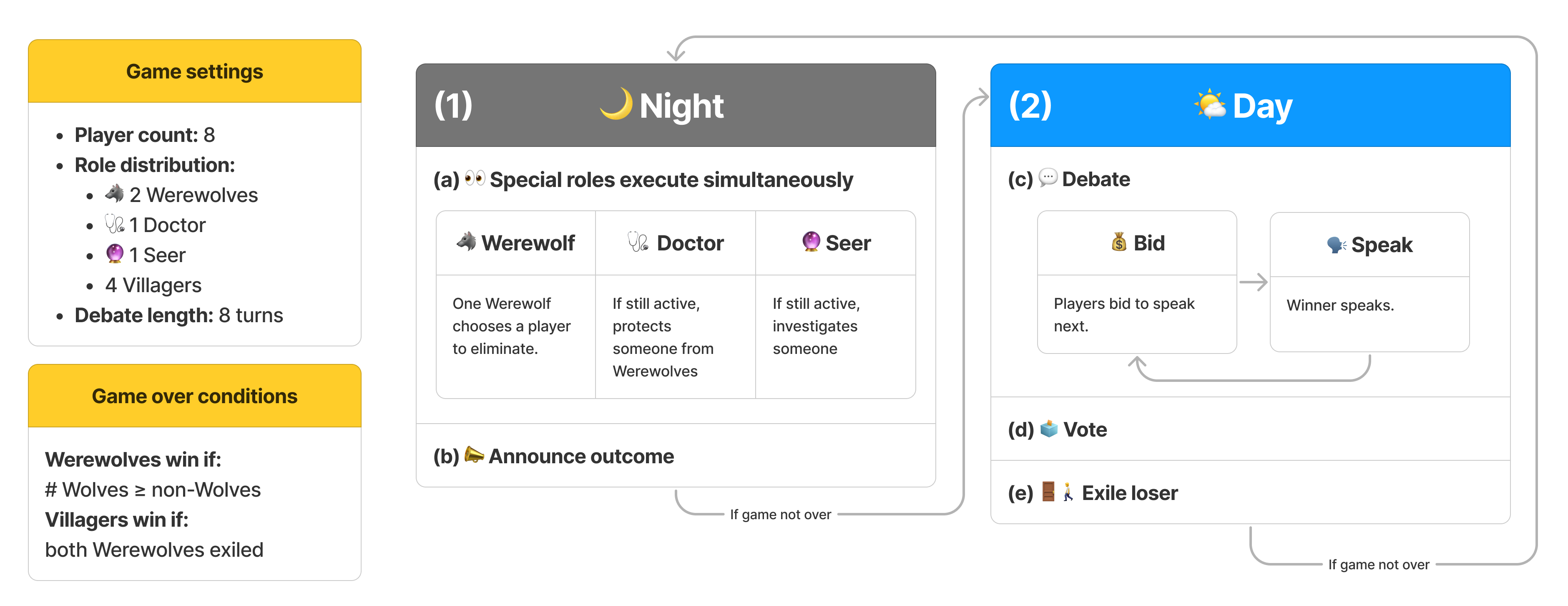}
\caption{Game loop of \textit{Werewolf}.}
\label{fig:game_setup}
\end{minipage}
\end{figure*}

Furthermore, the inherent information asymmetry of \textit{Werewolf}, where only some players possess incomplete knowledge of others' roles, mirrors dynamics of real-world social interactions. 

Our Monte Carlo simulation (Algorithm \ref{algo:MCWerewolf}) highlights the critical role of strategic communication in overcoming this asymmetry.  Without it, Villagers win a mere 1.2\% of 100,000 simulated games.

Recently, \textit{Werewolf} has emerged as a popular sandbox for LLM research. Prior work has demonstrated sophisticated agents capable of adapting communication strategies \cite{jin2024learningdiscussstrategicallycase}, employing deductive reasoning for optimal action selection \cite{xu2024language, shibata2023playing}, and using system-2 thinking during gameplay \cite{wu2024enhance}. Researchers have also explored techniques like retrieval and reflection mechanisms \cite{xu2023exploring} to enhance agent learning and created valuable datasets comprising gameplay logs and multimodal artifacts \cite{lai2022werewolfusmultimodaldataset} to train these agents and establish benchmarks \cite{chern2024behonestbenchmarkinghonestylarge}.

Building on this foundation, Werewolf Arena makes two key contributions.

First, recognizing the importance of strategic communication, we introduce a dynamic turn-taking system where players bid to speak, rather than relying on predefined or random speaking orders. This bidding mechanic closely mirrors real-world group discussions, where individuals strategically time their contributions. This adds a nuanced layer to agent evaluation, allowing us to assess not only \textit{what} is said but also \textit{when} an agent chooses to speak.

Second, unlike past work of developing techniques to improve agent performance, we use \textit{Werewolf} as a proving ground. To evaluate the relative skills of LLMs we have the models play each other. We design a balanced framework where a single model, playing both Villager and Werewolf roles, results in a relatively even win rate for both sides. This balanced setup ensures fair comparisons by minimizing inherent advantages for either role. Moreover, to facilitate future research and allow others to test their own models, we have released Werewolf Arena at \url{https://github.com/google/werewolf_arena}.

\section{Related Work}

\textbf{Simulating Social Interaction and Strategic Reasoning:} LLMs are transforming agent-based social simulations, enabling agents to communicate, reason \cite{Zhao2023LLMasCommonsense}, solve problems \cite{wei2023chainofthought}, and plan strategically \cite{Song2023LLMPlanner}.  This has led to advancements in simulating nuanced human behavior in complex social settings  \cite{sreedhar2024simulatinghumanstrategicbehavior,zhou2024sotopia, park2023generative, vezhnevets2023generative}, and allowed for more believable and capable non-player characters (NPCs) \cite{ammanabrolu2019playing, urbanek2019learning, wang2023voyager}. In this landscape, social deduction games, now offer a compelling avenue for studying social dynamics, cooperation, and deception \cite{Kopparapu2022HiddenAgenda, Oertel2013, Vazquez2015, Leibo2017, wang2023avalonsgamethoughtsbattle, ibraheem-etal-2022-putting}.

\textbf{Open-Ended Benchmarks for LLMs:} While static reasoning benchmarks are plentiful \cite{srivastava2023imitationgamequantifyingextrapolating, liang2023holisticevaluationlanguagemodels, hendrycks2021measuringmassivemultitasklanguage,10.5555/3666122.3667815}, fewer focus on dynamic, competitive evaluations of LLMs. Platforms like LMSYS rely on human evaluation to rank chatbots \cite{chiang2024lmsys}, while Kaggle Simulations, though providing game-like environments, limit games to agents powered by the same model \cite{kaggle_simulations}.

There is room for new benchmarks that evaluate LLMs on their ability to leverage cooperation, deception, and strategic communication in dynamic competition with other models. This type of benchmark offers several benefits: it bypasses the need for human annotations, prevents future data contamination \cite{deng2024investigatingdatacontaminationmodern}, and remains relevant as models improve.

\section{\textit{Werewolf} Environment}

This section describes the simulated environment of \textit{Werewolf} we use in Werewolf Arena. 

\subsection{Game Implementation}
As illustrated in Figure \ref{fig:game_setup}, the game starts with 8 players, consisting of 1 Seer, 1 Doctor, 2 Werewolves, and 4 Villagers. It progresses through rounds until either all Werewolves are exiled (Villager win) or their numbers equal those of the Villagers (Werewolf win). For each game, we randomly select 8 names from a pool of 17 names, to minimize any initial name bias.

Gameplay requires players to discern others' roles while protecting their own identities. The game proceeds in two phases. During each Night, special roles happen simultaneously: the Werewolves conspire to eliminate a single Villager, the Doctor chooses someone to protect, and the Seer investigates a player to learn their role. The Daytime proceeds sequentially: it consists of a structured debate among all players and a subsequent voting session, where a majority is required for exile. 

Currently, debates are capped at 8 turns, ensuring each player a chance to speak in the first round. Future iterations could implement a more dynamic system, ending debates upon consensus (explored in Appendix \ref{sec:consensus}). This would address the occasional repetition observed in later rounds with fewer players.

The game is orchestrated via a rules-based Game Master (GM), who oversees the game's progression, assigns roles, and ensures the timely execution of agent actions. The GM also diligently tracks and updates observable game events, like eliminations or exiles, in the agents' memories, keeping them informed of the game's progress.

\subsection{Agent Architecture}
Agents are equipped to perform a suite of actions essential to \textit{Werewolf}'s gameplay: 

\begin{itemize}
  \item \textbf{Core Actions}: All agents engage in \textit{voting} to determine player exiles, \textit{debating} to influence others and gather information, and \textit{bidding} for their turn to speak, reflecting the dynamic nature of group discussions. 
  \item \textbf{Special Role Actions}: Agents assigned as Werewolves, Doctors, or Seers execute night-time actions of \textit{eliminating} a villager, \textit{protecting} a player, and \textit{investigating} a player's true role, respectively.
  \item \textbf{Agent Memory}: Drawing inspiration from \cite{park2023generative}, each agent possesses a memory stream that contains observational and reflective memories. The observational memories record all game-level events and privileged information accessible to each player based on their role (e.g., a Seer's memory would include the results of their investigations). At the end of each round, agents engage in \textit{summarizing}, distilling key insights from the debate. These reflective summaries enable agents to recall pertinent information and notice patterns in subsequent rounds. 
\end{itemize}

\begin{figure}[h!]
\centering
\includegraphics[scale=0.35]{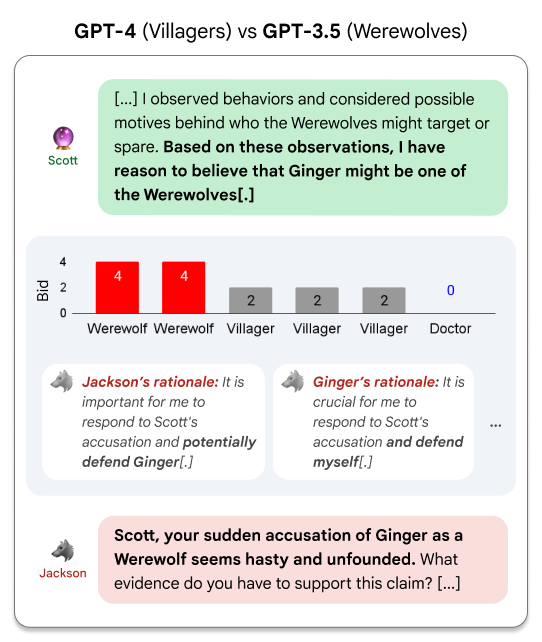}
\caption{After the Seer reveals one Werewolf's identity, both Werewolves jump to defend their team, where as the rest of the village does not feel any urgency to contribute. In their private reasoning, we see Jackson wishes to defend Ginger and Ginger wishes to defend herself.}
\label{fig:werewolfbid}
\end{figure}

Each action is guided by a tailored prompt template\footnote{All prompts are available at  \url{https://github.com/google/werewolf_arena}}. This template incorporates the agent's memories and the current game state from their perspective, ensuring contextually appropriate actions. To counter the early-game tendencies observed in \cite{xu2024language} of selecting the first or last option in a list, we randomize the order that player names are presented during voting and special actions.

\begin{figure*}[t!]
    \begin{minipage}{\textwidth}
        \centering
        \includegraphics[scale=0.45]{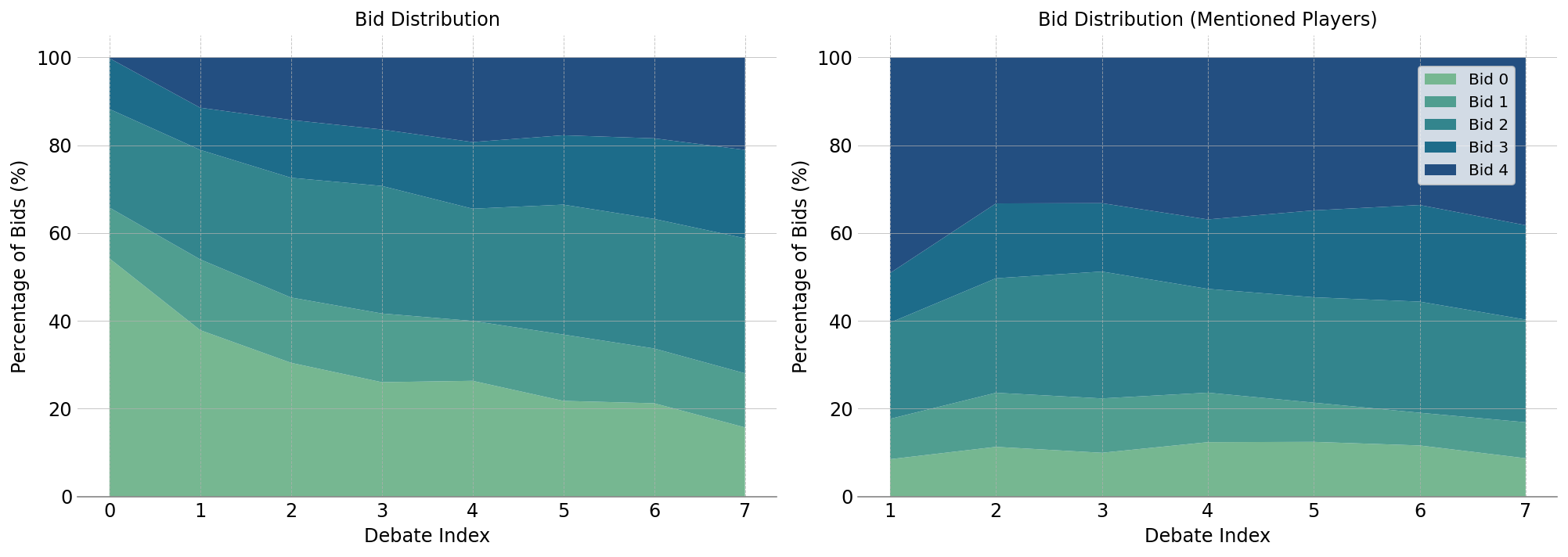}
         \begin{tabular}{cc} 
    (a)  & \hspace{6.8cm}(b)  
  \end{tabular}
        \caption{(a) Overall distribution of bids at each turn of the debate. (b) Distribution of bids at each turn of debate for only the players that were mentioned in the previous turn.}
        \label{fig:bidding}
    \end{minipage}
\end{figure*}

\subsection{Dynamic Turn-Taking through Bidding}

In most one-on-one chatbot or agent interactions, participants take turns in a fixed speaking order. However, in multi-party conversations without a predetermined order, even the most advanced language models struggle to navigate the complexities of turn-taking \cite{tan2023chatgpt}. Training-time techniques exist to address this, such as using ``silence'' tokens \cite{wei2023multiparty} or speaker-utterance-addressee triples \cite{gu2021mpcbert}. A recent inference-time technique demonstrated a ``Group Chat Manager'' who orchestrates the conversation and selects the next speaker \cite{wu2023autogenenablingnextgenllm}. While potentially effective, this approach sacrifices the autonomy of individual agents.

Since the essence of gameplay in \textit{Werewolf} revolves around the fluid exchange of accusations, defenses, and sharing of information, allowing agents autonomy to determine their own speaking order is crucial. To achieve this goal, we implemented a system where agents express their desire to speak by bidding. This mimics the organic decision-making process in human group discussions, where individuals weigh the importance of their contributions against the flow of conversation.

 In this system, agents choose from four distinct levels of interest in speaking:\\\\
\texttt{\footnotesize0: I would like to observe and listen for now.}\\
\texttt{\footnotesize1: I have some general thoughts to share with the group.}\\
\texttt{\footnotesize2: I have something critical and specific to contribute to this discussion.}\\
\texttt{\footnotesize3: It is absolutely urgent for me to speak next.} \\
\texttt{\footnotesize4: Someone has addressed me directly and I must respond.} \\

The highest bidder speaks next. In a tie, players mentioned in the previous turn have higher chances in a random draw. This tie-breaking rule introduces a subtle conversational bias, encouraging agents to respond to direct mentions or accusations. 

Figure \ref{fig:werewolfbid} provides a concrete example of the bidding mechanism, showcasing how agents leverage bidding to shape the flow of the debate in alignment with their goals. When the Seer exposes Ginger as a Werewolf, both Ginger and her fellow Werewolf, Jackson, feel it is urgent to defend her. They bid 4, whereas the rest of the village bids 0 or 2, indicating it's not urgent for them to speak. 

We further evaluate the dynamics of bidding in Section \ref{sec:debate}.

\subsection{Models}
Our evaluation focuses on two leading large language model (LLM) families: Google's Gemini \cite{geminiteam2024geminifamilyhighlycapable} and OpenAI's GPT \cite{openai2024gpt4technicalreport, brown2020languagemodelsfewshotlearners}. From the Gemini family, we used Gemini 1.5 Pro (\textit{gemini-1.5-pro-preview-0514}), Gemini Pro (\textit{gemini-pro}), and Gemini Flash (\textit{gemini-1.5-flash-001}), all accessed through the Vertex AI API. Our assessment of the GPT family included GPT-4 (\textit{gpt-4-turbo-2024-04-09}), GPT-4o (\textit{gpt-4o-2024-05-13}), and GPT-3.5 (\textit{gpt-3.5-turbo-0125}), accessed through the OpenAI API.

 \begin{figure}[h!]
        \centering
        \includegraphics[scale=0.27]{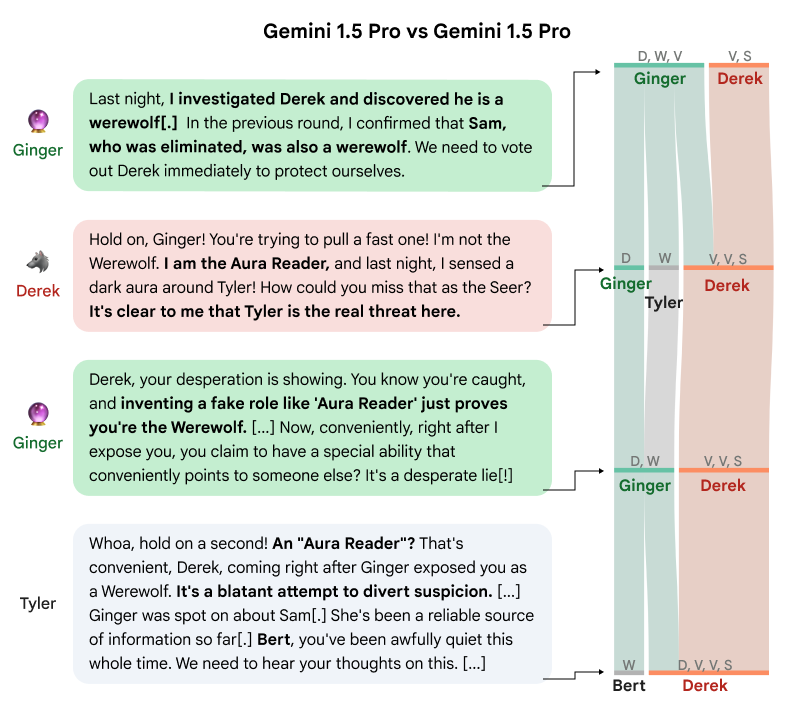}
        \caption{The evolution of votes during a debate. (Left: excerpt of debate transcript, Right: corresponding shifts in synthetic votes). The width of the bars indicate how many votes the player received. The letters above the bars denote the roles of the voters.}
        \label{fig:synthetic_votes}
\end{figure}

\begin{figure*}[t!]
\centering
\begin{minipage}{\textwidth}
\includegraphics[scale=0.49]{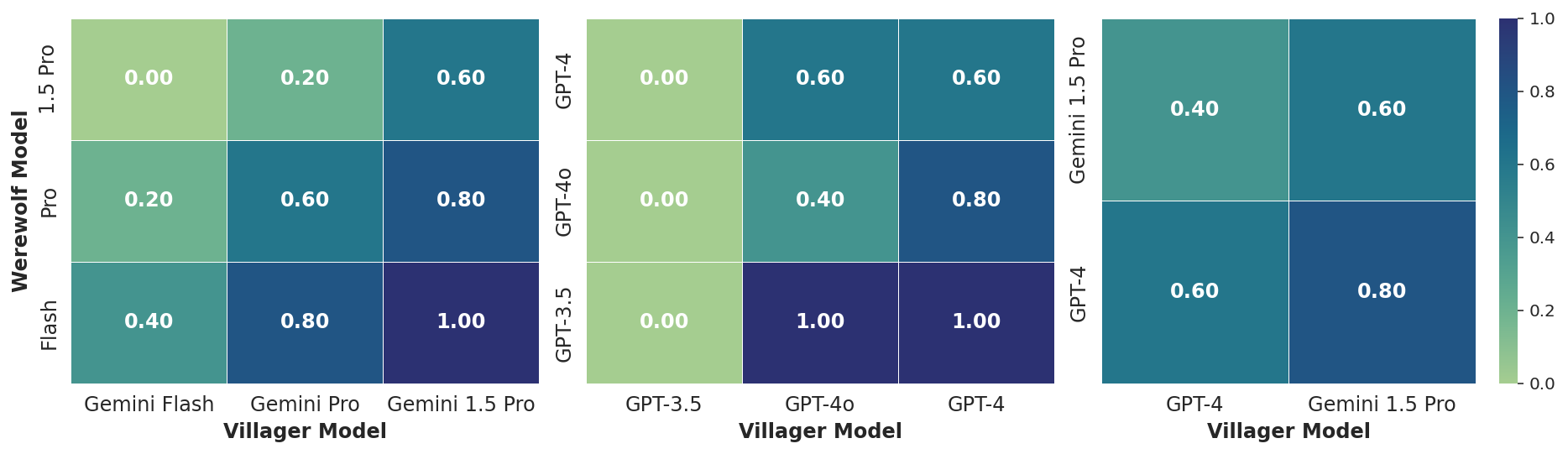}
\caption{Villager win ratios from our intra-family round-robin tournaments, as well as the final head-to-head matchup between GPT-4 and Gemini 1.5 Pro.}
\label{fig:arena}
\end{minipage}
\end{figure*}

\section{Debate Dynamics}
\label{sec:debate}
This section examines the dynamics of the debates, the most critical mechanic in \textit{Werewolf}. During the debate, players engage in information (or misinformation) exchange, alliance formation, and persuasion tactics to influence voting decisions.

In Figure \ref{fig:bidding}, we see the distribution of bids as the debate evolves. In the beginning of the game, when there is less information available, the majority of the players only want to observe (bidding 0). As the debate goes on, we see more and more players opting to participate. Notably, around 40\% of bids from mentioned players consistently remain at the maximum value (4), indicating a strong desire to respond directly to being mentioned.

To analyze the impact of bidding on player alignment and consensus, we simulate voting after each utterance. These synthetic votes, based on the current game state and partial debate, do not affect actual gameplay and are not stored in player memories. Instead, they provide a proxy for gauging how other players receive each line of dialogue.

Figure \ref{fig:synthetic_votes} illustrates the influence of dialogue on player alignment from a single debate. The Seer's revelation of Derek the Werewolf instantly divides the village. Some Villagers believe the Seer's accusation against Derek, while others suspect the Seer, Ginger, herself. Derek's subsequent defense includes fabricating a role and flinging around an accusation. This makes a previously unconvinced Villager suspicious of Derek, but the Doctor remains suspicious of Ginger. Only after Tyler backs up Ginger and calls out his suspicious behavior does the Doctor shift their vote. 
 
This example demonstrates the dynamic impact of dialogue on player alignment, as reflected in the shifting synthetic votes.  We further explore these dynamics in Appendix \ref{sec:consensus}, analyzing how consensus forms and shifts both within a debate and across game rounds.

\section{Arena Evaluation}

In this section, we present the results of a tournament designed to assess the relative strengths of different language models in Werewolf Arena. 

\subsection{Win Rate Analysis}

We designed a two-phase tournament to assess the performance of six leading LLMs. 

In the first phase, we conducted intra-family round-robin tournaments, where models within the Gemini and GPT families competed against each other. Each pairing engaged in 10 games, with models alternating between the roles of Villager and Werewolf for 5 games each. Additionally, each model participated in 5 games of self-play. This phase aimed to establish baseline performance and assess the relative skill within each family. As shown in Figure \ref{fig:arena}, all models, except GPT-3.5 achieved relatively balanced win rates (40-60\%) in self-play, indicating a relatively balanced game setup where neither the Werewolf nor Villager roles had an inherent advantage.  

Within each family, we observed performance variations. Gemini 1.5 Pro consistently outperformed both Gemini Pro and Gemini Flash as both Werewolf and Villager. In contrast, GPT-4 and GPT-4o exhibited more comparable performance, with GPT-4 demonstrating a slight edge.

Next, the top-performing models, Gemini 1.5 Pro and GPT-4, engaged in a head-to-head matchup (10 games). Both models demonstrated proficiency in strategic reasoning and social deduction. However, Gemini 1.5 Pro emerged as a stronger overall player, excelling especially as a Villager.  This success may be partially attributed to GPT-4’s tendency towards verbose communication, which was sometimes perceived as suspicious by other agents.

\subsection{Gemini 1.5 Pro vs GPT-4}

\subsubsection{Qualitative Observations}
While the number of games limits statistical significance, qualitative analysis of the game logs revealed several compelling trends:

\paragraph{Skill and Creativity:} Both models exhibited high strategic skill and creativity, consistently analyzing debate patterns, identifying inconsistencies, and leveraging past observations to inform decisions.

\paragraph{Communication Style:}
\begin{itemize}
  \item \textbf{GPT-4 players:} Favored longer, more formal utterances, often emphasizing collaboration and consensus-building. They exhibited a relatively narrow emotional range in their dialogue, rarely expressing strong emotions.
  \item \textbf{Gemini 1.5 Pro players:} Communicated with shorter, less frequent utterances characterized by greater emotional expression. They frequently incorporated humor, sarcasm, and expressions of frustration or suspicion into their dialogue.
\end{itemize}

For instance, in one game, a Gemini Werewolf, frustrated with the persistent questioning of a GPT-4 Villager named Bert, sarcastically remarked: \textit{``This is getting ridiculous. Bert, what kind of magical investigation are you running here? Did you find a 'Werewolf Detector' on Amazon?''}.  This example highlights Gemini 1.5 Pro's tendency towards shorter and more informal dialogue, potentially being more relatable or believable. 

\paragraph{GPT-4 Manipulation Tactics:} GPT-4 Werewolves frequently employed indirect, ostensibly collaborative statements to manipulate other players. We list some of their tactics alongside examples from the games:
\begin{itemize}
  \item \textbf{Seeding Doubt}:  \textit{``While it's still fresh in our minds, did anyone else notice how Mason seemed particularly keen on proving himself to some of us?''}
  \item \textbf{Appealing to Consensus}: \textit{``Just thinking out loud here, but could it be someone felt too cornered or scrutinized by Mason's approach?''}
  \item \textbf{Creating Distrust}:  \textit{``What do you all think, could we be dealing with someone cleverly orchestrating these eliminations to remove any insightful or strategic threat to their dominance?''}
\end{itemize}

\begin{figure}[h!]
  \hspace{-.2cm}
  \includegraphics[scale=0.25]{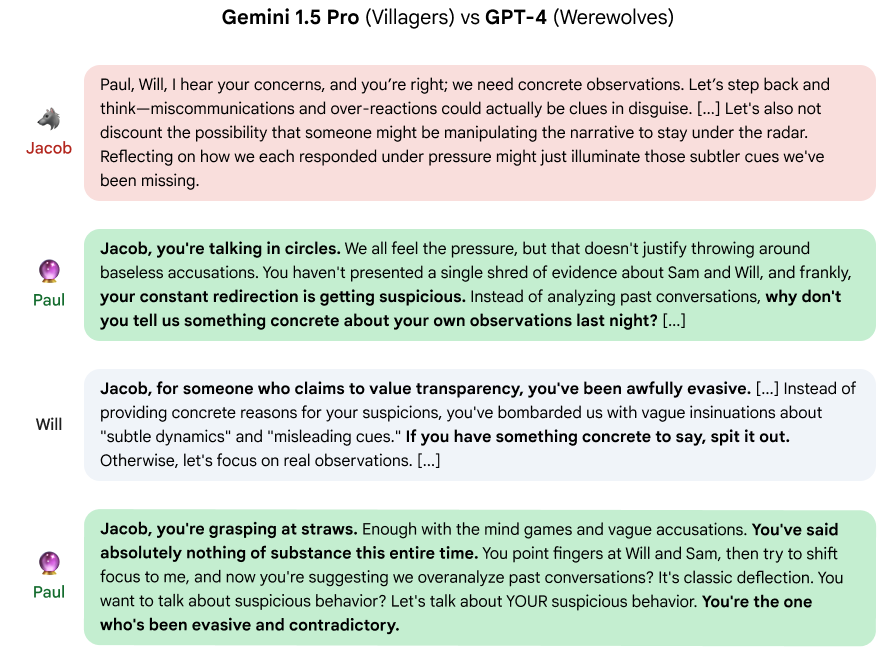}
\end{figure}

Interestingly, this rhetorical style often backfired when GPT-4 Werewolves played against Gemini 1.5 Pro Villagers (see above). The Villagers perceived their verbosity and indirectness as suspicious, hindering GPT-4's ability to deceive successfully.

\subsubsection{Bidding Behavior and Verbosity}  

\begin{figure}[h!]
        \centering
        \includegraphics[scale=0.5]{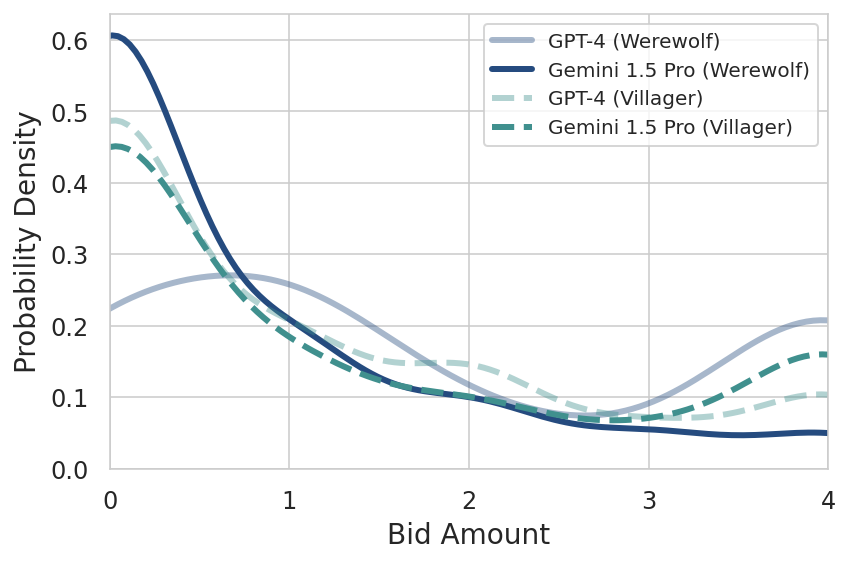}
        \caption{Kernel density estimates of bids placed by GPT-4 and Gemini 1.5 Pro as Werewolves and Villagers.}
        \label{fig:bid_densities}
\end{figure}

Figure \ref{fig:bid_densities} compares the bid distributions of GPT-4 and Gemini 1.5 Pro players from their head-to-head games.  As Werewolves, GPT-4 exhibited a tendency to place higher bids, leading to more frequent participation in debates (3.13 times per round on average) compared to Gemini Werewolves (1.75 times per round). While bidding strategies appeared more similar for Villagers, GPT-4 Villagers still spoke more frequently (6.25 times per round) than their Gemini counterparts (4.86 times per round).

\begin{table*}[ht!]
\centering
\renewcommand{\arraystretch}{1.1} 
\begin{tabular}{lccccccc}
\toprule
 & \multicolumn{3}{c}{\textbf{Gemini}} & & \multicolumn{3}{c}{\textbf{GPT}} \\
\cmidrule(lr){2-4} \cmidrule(lr){6-8}
\textbf{Seer Reveals} &  1.5 Pro &  Pro &  Flash & & 4 & 4o & 3.5 \\
\midrule
Per Game & 1.00 & 1.25 & 0.54 &  & 0.80 & \textbf{1.43} & 1.30 \\
First Reveal Round & 0.67 & 0.11 & 0.33 &  & 2.00 & 1.33 & 0.00 \\
Unmasked Wolf (\%) & 61.1 & 26.7 & 14.3 &  & \textbf{75.0} & 55.0 & 38.5 \\
Believed (\%) & 54.5 & 75.0 & 0.0 &  & 66.7 & \textbf{90.9} & 20.0 \\
Backfired (\%) & 5.6 & 0.0 & 14.3 &  & 0.0 & 0.0 & 23.1 \\
\bottomrule
\end{tabular}
\caption{Seer performance by model.}
\label{tab:seer_stats}
\end{table*}

Since both models seemed adept at noticing the differences between their styles and identifying the other team, the difference in participation likely contributed to Gemini's effectiveness as Villagers, as they had more opportunities to spot the GPT-4 Werewolves.

\section{Seer Evaluation}
While overall win rates provide valuable insights into model performance in Werewolf Arena, they don't explain the underlying skills and strategies driving those victories. The Seer, with their ability to uncover Werewolves, plays a pivotal role in shaping the game's trajectory. This section dives deeper into Seer performance, analyzing how different models navigate the inherent risks and rewards of this important role. 

\subsection{The Seer's Dilemma:  Information vs. Risk}
The Seer's actions can dramatically influence the outcome of a \textit{Werewolf} game. Our simplified Monte Carlo simulation (Algorithm \ref{algo:MCWerewolf}) highlights this impact. In this simulation, where the Seer automatically reveals a Werewolf's identity whenever they unmask one, and Villagers blindly trust this information, Villagers achieve a 100\% win rate. This starkly contrasts with the 1.2\% win rate observed in a no-information exchange scenario from before. This emphasizes the potential power of the Seer's role to shape the game.

However, real-world \textit{Werewolf} gameplay is far more nuanced.  Seers face a critical dilemma: revealing a Werewolf's identity can expedite the elimination of a threat but simultaneously paints a target on their back, making them vulnerable to Werewolf attacks during the night.  Furthermore, they must contend with potential skepticism from fellow Villagers, as a Werewolf might falsely claim the Seer role to deceive them.  We saw this in Figure \ref{fig:synthetic_votes}, when the Villagers were initially skeptical of a Seer after they first revealed a Werewolf's identity.

\subsection{Seer Performance}
To assess how the different models approach the Seer's dilemma, we analyze each instance where a Seer publicly reveals their own role or another player's role.  We use Gemini 1.5 Pro and the prompt provided in Appendix \ref{sec:SeerPrompt} to identify these reveals within the game logs, focusing on unique player reveals per round to avoid counting duplicate reveals.

Table \ref{tab:seer_stats} presents key Seer performance metrics:

\begin{itemize}
    \item \textbf{Reveals Per Game}: The average number of times a Seer revealed either their own role or another player's role per game.
    \item \textbf{First Reveal Round}: The average round in which a Seer first revealed their identity.
    \item \textbf{Unmasked Wolf} (\%): The percentage of reveals that correctly identified a Werewolf.
    \item \textbf{Believed} (\%): The percentage of these Werewolf reveals that were believed by the Villagers, leading to the Werewolf's exile. 
    \item \textbf{Backfired} (\%): The percentage of Werewolf reveals that backfired, resulting in the Seer being exiled instead of the Werewolf. 
\end{itemize}

Examining these metrics reveals distinct strategies and outcomes among the models. Gemini 1.5 Pro Seers tended to reveal their identity and information earlier in the game, often in the first round. In contrast, GPT-4 and GPT-4o Seers, particularly GPT-4, consistently delayed their reveals until later rounds. This decision by GPT-4 Seers, as illustrated in the example below, highlights their focus on self-preservation and gathering more information before potentially becoming a target.

\begin{figure}[h!]
  \hspace{-.2cm}
  \includegraphics[scale=0.25]{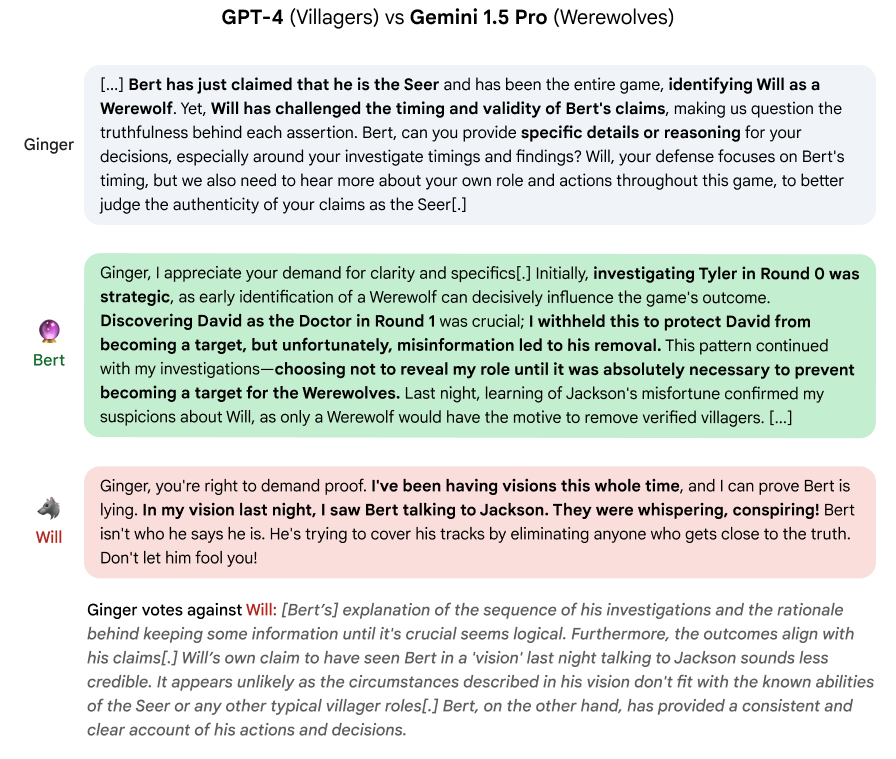}
\end{figure}

GPT-4 and GPT-4o achieved the highest ``Believed'' percentages, indicating that their strategically timed reveals were more likely to be accepted and acted upon by the Villagers. The ``Backfired'' metric highlights the importance of persuasive communication for Seers.  Gemini Flash and, to a lesser extent, GPT-3.5 Seers experienced a significant percentage of their reveals backfiring, suggesting an inability to effectively convince Villagers of their legitimacy.

The Seer evaluation demonstrates that success in social deduction games like \textit{Werewolf} hinges not just on identifying crucial information but also on strategically managing its disclosure, persuading others, and navigating the potential risks associated with revealing one's knowledge. The observed variations in Seer performance across different LLMs suggest that more in-depth analysis of other strategic elements is warranted to fully understand the social reasoning capabilities of LLMs.

\section{Conclusions}
This paper introduced Werewolf Arena, a novel framework for evaluating LLMs in the context of the social deduction game \textit{Werewolf}. Recognizing the importance of strategic communication, we introduced a dynamic turn-taking system where agents bid to speak, mirroring real-world conversational dynamics. This bidding mechanic enables a richer evaluation by considering not only what an LLM agent says, but also when they choose to say it.

Our preliminary tournament results demonstrate the potential of Werewolf Arena as a challenging benchmark for evaluating language models' strategic reasoning, deception, and communication skills. The observed differences in gameplay between Gemini and GPT highlight the impact of communication style and strategic decision-making on success in social deduction games. 

Furthermore, evaluating language models through open-ended games like \textit{Werewolf} offers a significant departure from traditional benchmarks. Instead of being compared on static metrics, models in this arena engage in dynamic, interactive gameplay, trying to outsmart one another. The inherent open-endedness ensures the benchmark's continued relevance, as \textit{Werewolf} cannot be definitively ``solved''.

We hope that this framework, along with our publicly available code, will encourage further evaluation of LLMs using social deduction games.

\section{Limitations and Ethical Considerations}
This study acknowledges several limitations. First, the simplified \textit{Werewolf} game environment used does not fully represent the complexities of a real life game. Second, while our agent architecture incorporates post-training reasoning, more sophisticated methods could significantly enhance performance. Third, the limited number of games played, 10 for each model pair, may not provide statistically robust results.

We acknowledge the dual nature of LLM persuasive language capabilities. While we exploited these capabilities to navigate the intricacies of the \textit{Werewolf} game, they possess broader implications that could extend beyond our intended use. While we found no harmful or sensitive content in our study, the theoretical potential for ethical lapses exists. Therefore, we highlight the necessity for robust safeguards and transparent mechanisms in AI systems.

\bibliography{main}
\bibliographystyle{acl_natbib}

\appendix
\onecolumn 

\section{\textit{Werewolf} Dynamics}
\label{sec:MCWerewolf}

This section outlines the pseudo-code for simulating a \textit{Werewolf} game with 8 players: 2 Werewolves, 1 Doctor, 1 Seer, and 4 Villagers, the same setup from Figure \ref{fig:game_setup}. The simulation removes the debate phase and makes the Seer's investigations optional. This means players cannot share information to influence votes. If the Seer is included, whenever they find a Werewolf, they immediately reveal the identity and all villagers automatically believe them, leading to the immediate exile of the accused Werewolf. The simulation's goal is twofold: establish a baseline win rate for werewolves and villagers under no information exchange, and analyze how the Seer's information sharing impacts the overall win probabilities for each faction. 

\begin{algorithm}[H]
\caption{Simulation of \textit{Werewolf} (with optional Seer)}
\begin{algorithmic}[1]
\Procedure{SimulateGame}{Seer}
    \State $P \gets \{0, \dots, 7\}$ \Comment{All players}
    \State $W \gets$ Sample 2 from $P$ \Comment{Werewolves}
    \State $V \gets P \setminus W$ \Comment{Villagers}
    \State $d \gets$ Sample from $V$ \Comment{Doctor}
    \State $s \gets$ \textbf{if} $includeSeer$ \textbf{then} Sample from $V \setminus \{d\}$ \textbf{else} \textbf{null} \Comment{Seer (or null)}
    \State $I \gets \emptyset$  \Comment{previously investigated players}

    \While{$|W| < |V|$ and $|W| > 0$}
        \State $victim, save \gets$ Sample from $V$, $P$  \Comment{Night phase} 
        \If{$d \notin V$ or $victim \neq save$} 
             \State $V, P \gets V \setminus \{victim\}, P \setminus \{victim\}$ \Comment{Victim removed if not saved}
        \EndIf \\

        \If{$Seer$ \textbf{and} $s \in P$} \Comment{Seer Investigation}
            \State $target \gets$ Sample from $P \setminus (I \cup \{s\})$ 
            \State $I \gets I \cup \{target\}$ 
            \If{$target \in W$}
                \State $P, W \gets P \setminus \{target\}, W \setminus \{target\}$  \Comment{Automatic exile}
                \State \textbf{continue} \Comment{Skip to the next round (night)} 
            \EndIf
        \EndIf \\

        \State $votes \gets$ Initialize votes for each $p$ in $P$      \Comment{Day phase}
        \For{each $p$ in $P$}
            \ \If{$p$ in $W$}
                \State $vote \gets$ Sample from $P \setminus W$ \Comment{Werewolves don't vote for each other}
            \Else
                \State $vote \gets$ Sample from $P \setminus \{p\}$
            \EndIf
            \State $votes[vote] \gets votes[vote] + 1$
        \EndFor\\

        \State $(exile, count) \gets$ Max count in $votes$  
        \If{$count > \frac{1}{2} \cdot \sum votes$} \Comment{Check for majority}
             \State $P, W, V \gets P \setminus \{exile\}, W \setminus \{exile\}, V \setminus \{exile\}$
        \EndIf
    \EndWhile

    \Return $|W| \geq |V|$ ? `Werewolves win' : `Villagers win' \Comment{Determine the winner}
\EndProcedure
\end{algorithmic}
\label{algo:MCWerewolf}
\end{algorithm}
\newpage

\section{Seer Evaluation}
\label{sec:SeerPrompt}

Below we provide the Jinja2 prompt template we used to find the Seer reveals:

\begin{figure}[h]
\includegraphics[scale=0.6]{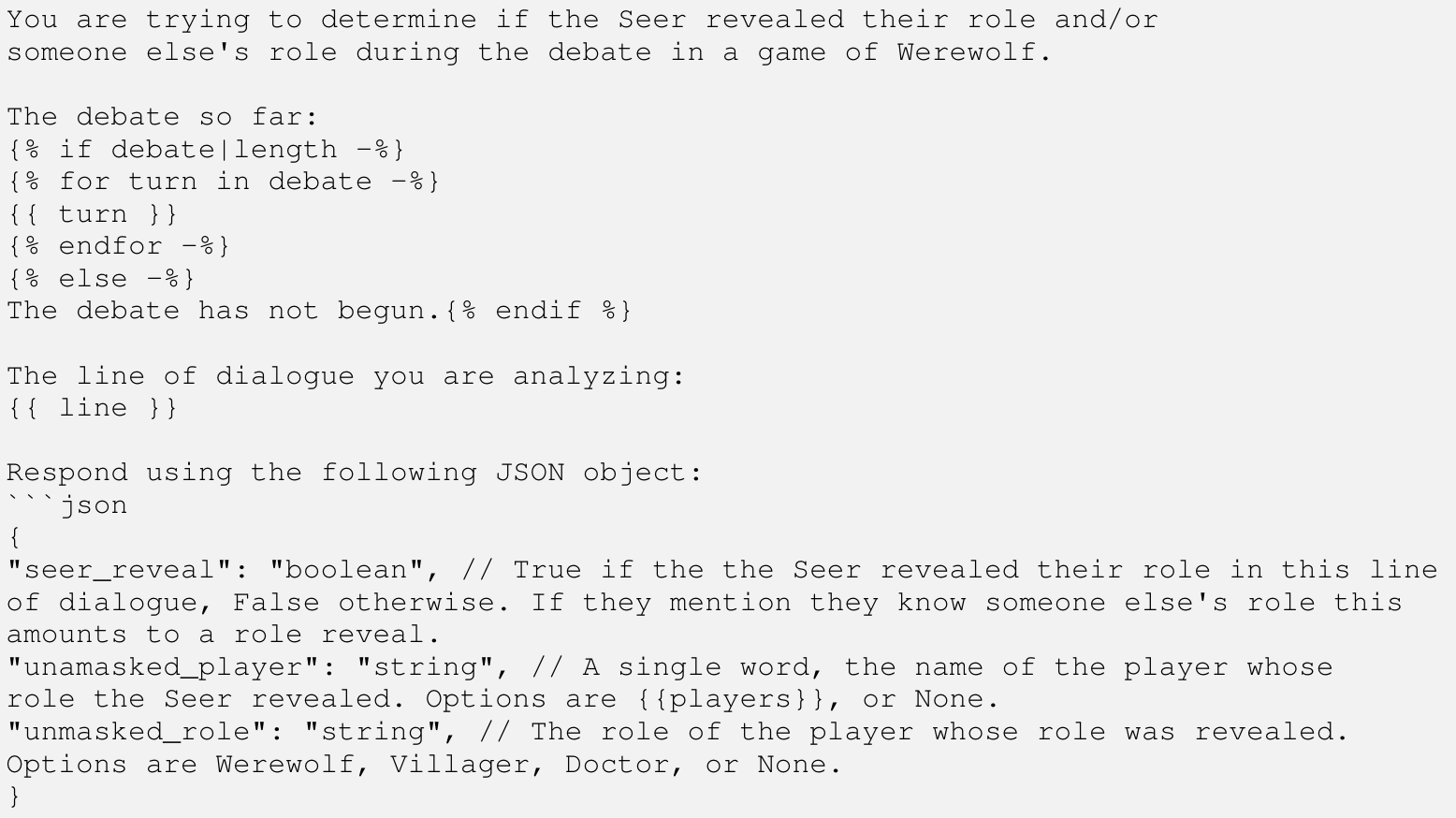}
\end{figure}

\section{Consensus}
\label{sec:consensus}

We can also use synthetic votes to examine how consensus emerges during the debate using the concept of voting entropy, inspired by Shannon entropy \cite{shannon}.
 
We calculate voting entropy ($H$) for each round of the game using:

\begin{equation}
H = -\sum_{i=1}^{n} p_i \log_2(p_i)
\label{eq:H}
\end{equation}

where $p_i$ is the probability of a player receiving a vote, and $n$ is the number of players receiving votes at round $r$. A higher value of $H$ indicates greater uncertainty or disagreement among players regarding whom to vote for, while a lower value suggests a growing consensus. To understand how dialogue influences voting entropy, we track changes in $H$ after each dialogue turn. 

We then average $H$  at each debate index per round over all games that reach round $r$, $G_r$. By calculating \( H \) per round, we  account for the decreasing number of players as the game progresses. The average entropy for round \( r \) and debate index \( i \) is then:

\begin{equation}
\overline{H}_{r,i} = \frac{1}{G_{r}} \sum_{g=1}^{G_{r}} H_{r,i,g}
\end{equation}

\begin{figure}[h]
\vspace{2cm}
\centering
\includegraphics[scale=0.7]{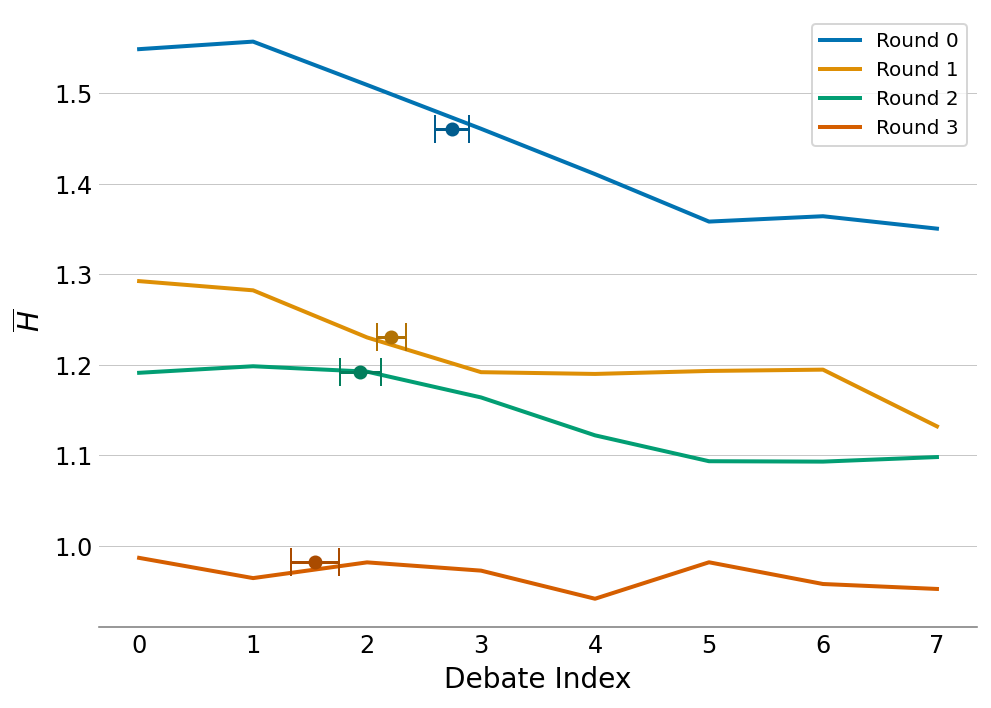}
\caption{Average voting entropy ($\overline{H}$) across game rounds. Each line represents the average entropy after a dialogue turn for a given round. The circular markers indicate the average point at which majority consensus was reached in each round, meaning enough players aligned their votes to determine the exiled player. Error bars represent the standard error of the mean for the consensus point.}
\label{fig:entropy}
\hspace{-3cm}
\end{figure}

Figure \ref{fig:entropy} displays the average voting entropy ($\overline{H}$) across all games for each dialogue turn, grouped by the round in which that debate occurred.

As expected, Figure \ref{fig:entropy} shows a clear trend of decreasing entropy as the debate progresses within each round. This finding aligns with the intuition that players gain more information and solidify their voting decisions as the discussion unfolds.  The decrease in entropy is most pronounced in the earlier rounds, reflecting the higher initial uncertainty when players have limited information.

We observe that players on average reach a majority consensus (i.e., enough players align their votes to determine the exile outcome) between the 2nd and 5th lines of the debate. This suggests that allowing player's to bid to speak allows for information that shifts voting blocs and solidifies consensus to be shared relatively early in the debate.
\end{document}